\documentclass[5p]{elsarticle}
\usepackage[pagebackref=true,breaklinks=true,letterpaper=true,colorlinks,bookmarks=false]{hyperref}

\usepackage{caption}
\journal{NeuroComputing}

\usepackage{amsmath,amssymb,multirow}
\usepackage{graphicx}
\usepackage{bm}
\usepackage{algorithm}
\usepackage{algorithmic}
\usepackage{subfigure,color}

\begin{document}

\begin{frontmatter}

\title{Fast and Robust Dynamic Hand Gesture Recognition via \\ Key Frames Extraction and Feature Fusion}

\author{Hao Tang$^{1}$, Hong Liu$^{2*}$, Wei Xiao$^3$, Nicu Sebe$^1$}
\cortext[mycorrespondingauthor]{Corresponding author.}

\address{
$^1$Department of Information Engineering and Computer Science, University of Trento, Trento, Italy \\
$^2$Key Laboratory of Machine Perception, Shenzhen Graduate School, Peking University, Beijing, China \\
$^3$Lingxi Artificial Intelligence Co., Ltd, Shen Zhen, China\\  
}
\tnotetext[label2]{E-mail: \{hao.tang, niculae.sebe\}@unitn.it; hongliu@pku.edu.cn; xiaoweithu@163.com; 
}

\begin{abstract}
	Gesture recognition is a hot topic in computer vision and pattern recognition, which plays a vitally important role in natural human-computer interface.
	Although great progress has been made recently, fast and robust hand gesture recognition remains an open problem,
	since the existing methods have not well balanced the performance and the efficiency simultaneously.
	To bridge it, this work combines image entropy and density clustering to exploit the key frames from hand gesture video for further feature extraction, which can improve the efficiency of recognition.
	Moreover, a feature fusion strategy is also proposed to further improve feature representation, which elevates the performance of recognition.
	To validate our approach in a ``wild'' environment, we also introduce two new datasets called HandGesture and Action3D datasets.
	Experiments consistently demonstrate that our strategy achieves competitive results on Northwestern University, Cambridge, HandGesture and Action3D hand gesture datasets.
	Our code and datasets will release at \url{https://github.com/Ha0Tang/HandGestureRecognition}.
\end{abstract}

\begin{keyword}
Hand gesture recognition; Key frames extraction; Feature fusion; Fast; Robust.
\end{keyword}

\end{frontmatter}


\section{Introduction}
\label{sec:intro}
Gesture recognition is to recognize category labels from an image or a video which contains gestures made by the user.
Gestures are expressive, meaningful body motions involving physical movements of the fingers, hands, arms, head, face, or body with the intent of:
conveying meaningful information or interacting with the environment.

Hand gesture is one of the most expressive, natural and common type of body language for conveying attitudes and emotions in human interactions.
For example, in a television control system, hand gesture has the following attributes: ``Pause'',``Play'', ``Next Channel'', ``Previous Channel'', ``Volume Up'', ``Volume Down'' and ``Menu Item''. While in a recommendation system, hand gesture can express ``Like'' or ``Dislike" emotions of users.
Thus, it is one of the most fundamental problems in computer vision and pattern recognition, and has a wide range of applications such as
virtual reality systems \cite{wang2015superpixel}, interactive gaming platforms \cite{ren2013robust},
recognizing sign language \cite{hikawa2015novel,marin2014hand,kuznetsova2013real},
enabling very young children to interact with computers \cite{yao2014contour},
controlling robot \cite{prasuhn2014hog,neto2013real}, 
practicing music conducting \cite{schramm2014dynamic},
television control \cite{lian2014automatic,freeman1995television},
automotive interfaces \cite{ohn2014hand,ohn2013power},
learning and teaching assistance \cite{sathayanarayana2014towards,sathyanarayana2013hand}, and hand gesture generation \cite{tang2018gesturegan}.

There has been significant progress in hand gesture recognition, however, some key problems e.g., fast and robust are still challenging.
Prior work usually puts emphasis on using whole data series, which always contain redundant information, resulting in degraded performance.
For examples, Wang et al. \cite{wang2015superpixel} present a superpixel-based hand gesture recognition system based on a novel superpixel earth mover's distance metric.
Ren et al. \cite{ren2013robust} focus on building a robust part-based hand gesture recognition system.
Hikawa and Kaida \cite{hikawa2015novel} propose a posture recognition system with a hybrid network.
Moreover, there are many approaches are also proposed for action or video recognition task, such as \cite{liu2013learning,yu2016structure,tang2015gender,tran2015learning,shao2016kernelized,wang2016temporal,ji20133d,liu2016sequential,qin2017zero,liu2015sdm}.
Liu and Shao \cite{liu2013learning} introduce an adaptive learning methodology to extract spatio-temporal features, simultaneously fusing the RGB and depth information, from RGB-D video data for visual recognition tasks.
Liu et al. \cite{liu2015sdm} propose to combine the Salient Depth Map (SDM) and the Binary Shape Map (BSM) for human action recognition task.
Simonyan et al. \cite{simonyan2014two} propose a two-stream ConvNet architecture which incorporates spatial and temporal networks to extract spatial and temporal features.
Feichtenhofer et al. \cite{feichtenhofer2016convolutional} study a number of ways of fusing ConvNet towers both spatially and temporally in order to best take advantage of this spatio-temporal information.
In sum, all these efforts endeavor to decrease the computation burden in each solo frame, while overlooking all processing schemes in the whole frames would incur more computation burden than a few selected representative frames, which is a fundamental way to decrease the computation burden, greatly.
This paper is devoted to bridge the gap between fast and robust hand gesture recognition, simply using solo popular cue e.g., RGB, which ensures great potential in practical use.

Key frames, also known as representative frames, extract the main content of a data series, which could greatly reduce the amount of processing data.
In \cite{zhao2008information}, the key frames of the video sequence are selected by their discriminative power and represented by the local motion features detected in them and integrated from their temporal neighbors.
Carlsson and Sullivan \cite{carlsson2001action} demonstrate that specific actions can be recognized in long video sequence by matching shape information extracted from individual frames to stored prototypes representing key frames of the action.
However, we regard every frame in a video as a point in the 2-D coordinate space.
Since we are focusing on distinguishing dynamic gesture from a data series while not reconstructing it, we simply introduce a measure to find which frames are more important for distinguishing and which are not.
In consideration of information entropy  \cite{brink1996using,min2013novel,wang2017ship} could be a useful measurement to quantify the information each frame contains, we introduce frame entropy as an quantitative feature to describe each frame and then map these values into a 2-D coordinate space.
How to describe this 2-D space is a hard nut to crack for its uneven distribution.
Therefore, we further propose an integrated strategy to extract key frames using local extreme points and density clustering.
Local extreme points includes the local maximum and local minimum points, which represent the most discriminative points of frame entropy.
Shao and Ji \cite{shao2009motion} also propose a key frame extraction method based on entropy.
However, the differences between \cite{shao2009motion}  and the proposed method are two-folder: (i) The entropy in \cite{shao2009motion}  is calculated on motion histograms of each frame, while the proposed method directly calculate on each frame.
(ii) \cite{shao2009motion}  simply to find peaks in the curve of entropy and use histogram intersection to output final key frames, while the proposed method first selects the local peaks of entropy and then use density clustering to calculate the cluster centers as the final key frames.
Density clustering \cite{rodriguez2014clustering} is the approach based on the local density of feature points, which is able to detect local clusters, while previous clustering approaches such as dynamic delaunay clustering \cite{kuanar2013video}, k-means clustering \cite{de2011vsumm}, spectral clustering  \cite{vazquez2013spatio} and graph clustering \cite{panda2014scalable} cannot detect local clusters due to the fact that they only rely on the distance between feature points to do clustering.

In order to promote the accuracy, we also present a novel feature fusion method that combines appearance and motion cues.
After extracting key frames, we replace the original video sequence with the key frames sequence, which could greatly enhance the time efficiency at the cost of accuracy.
This feature fusion strategy takes advantage of both the motion and the appearance information in the spatiotemporal activity context under the hierarchical model.
The experimental results show that
the method proposed is accurate and effective for dynamic hand gesture recognition on four datasets.
To summarize, the main contributions of this paper are:
\begin{itemize}
	\item A novel key frames extraction method is proposed, which improves efficiency of hand gesture processing.
	\item A feature strategy is presented in which appearance and motion cues are fused to elevate the accuracy of recognition.
	\item Experiments demonstrate that our method achieves the balance between efficiency and accuracy simultaneously in four datasets.
\end{itemize}

\section{Key Frames Extraction and Feature Fusion Strategy for Hand Gesture Recognition}
\label{sec:propo}
In this section, we will introduce the proposed key frames extraction and feature fusion strategy.
\subsection{Key Frames Extraction}
Key frames extraction is the key technology for video abstraction, which can remove the redundant information in the video greatly.
The algorithm for key frames extraction will affect the reconstruction of video content.
If a frame in video $V$ can be represented by $f_i$, where $i$ is $(1, 2, . . . , n)$ and $n$ is the total number of frames in video $V$.
Hence, the key frames set $S_\mathit{Keyframes}$ is defined as follows:
\begin{equation}
\label{equ:keyfra}
S_\mathit{Keyframes} = f_\mathit{Keyframes}(V),
\end{equation}
where $f_\mathit{Keyframes}$ denotes the key frames extraction procedure.

\begin{figure*}[!htbp]
	\begin{centering}
		\centering
		\setcounter{subfigure}{0}
		\subfigure[A hand gesture sequence sample from the Northwestern University hand gesture dataset, which contains 26 frames. The key frames obtained by our method are in green boxes, which are the 2, 9, 14, 20 and 26 frames.]  {\includegraphics[width=1\linewidth]{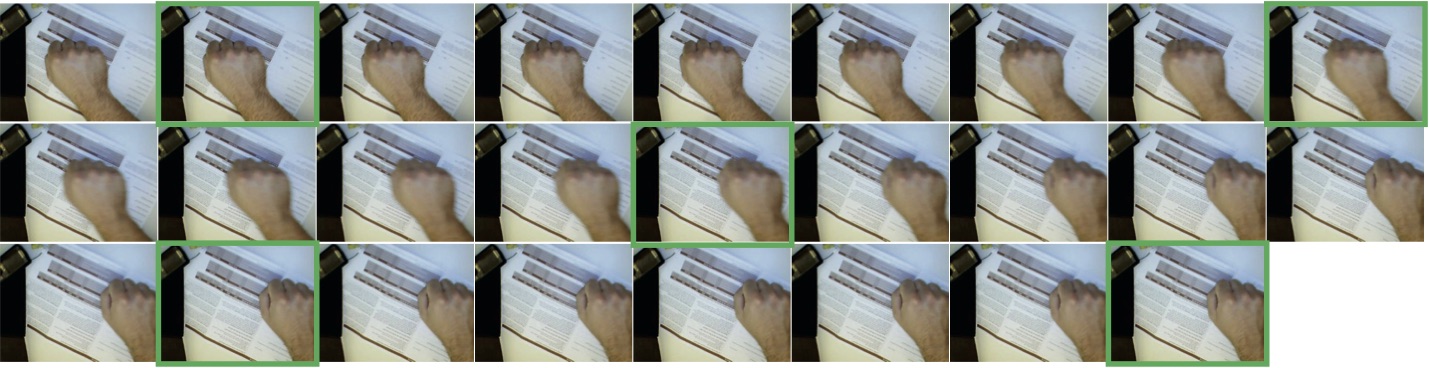}} \\
		\subfigure[Calculate the image entropy.]    {\includegraphics[height=1.9in]{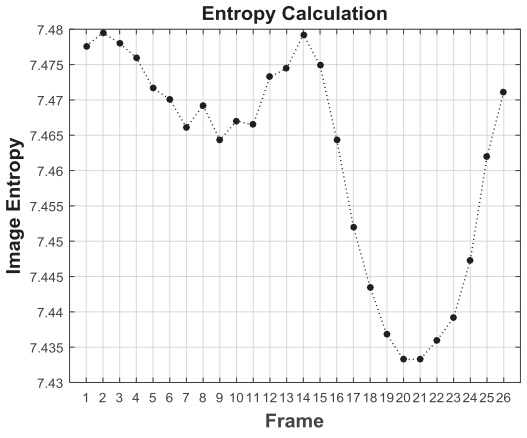}}
		\subfigure[Select the local peak points.]   {\includegraphics[height=1.9in]{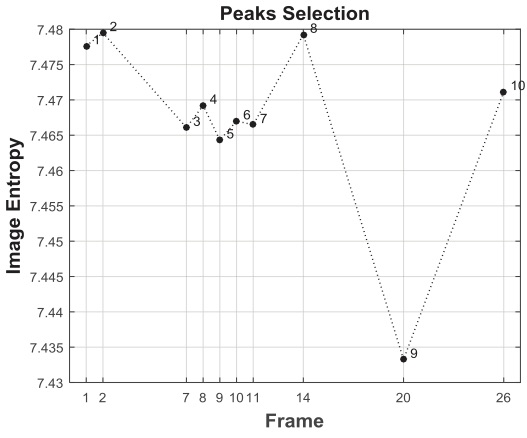}}
		\subfigure[Calculate $\rho$ and $\delta$.]  {\includegraphics[height=1.9in]{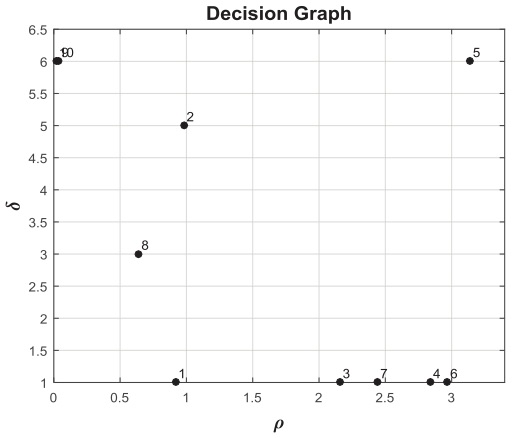}} \\
		\subfigure[Select the number of clustering. In this case, we choose 5 clusters.]  {\includegraphics[height=1.9in]{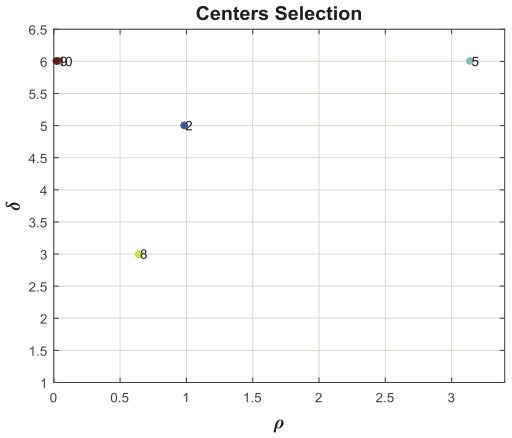}}
		\subfigure[The final results of clustering. The points of 2, 5, 8, 9 and 10 are the clustering centers, therefore, the corresponding frames (2, 9, 14, 20 and 26) are the key frames of original sequence.]{\includegraphics[height=1.9in]{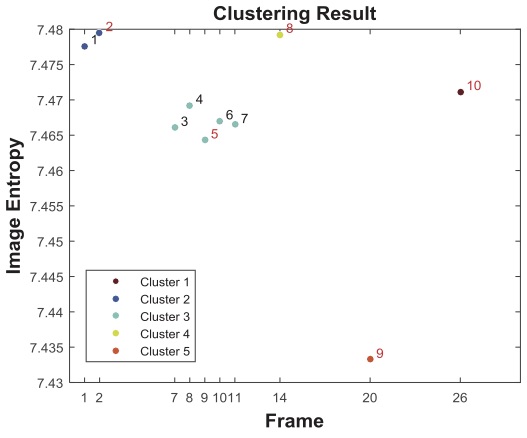}} \\
		\subfigure[The key frames are the 2, 9, 14, 20 and 26 frames. Now we use this sequence to replace the original sequences for the next step.]{\includegraphics[width=0.7\linewidth]{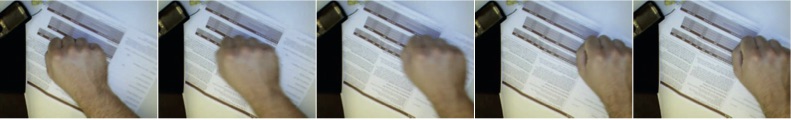}} \\
		\caption{The framework of the proposed key frames extraction method.}
		\label{Fig:key_frame_extra}
	\end{centering}
\end{figure*}

In this paper, a method of key frames extraction based on image entropy and density clustering is proposed, as we can see from Figure \ref{Fig:key_frame_extra}.
Our key frames extraction methods are mainly divided into three steps, namely, 1) calculating image entropy, 2) finding local extreme points and 3) executing density cluster.
The following section would expand upon on it.

\subsubsection{Image Entropy.}
In this section, we try to find a proper descriptive index to evaluate each frame in a video, facilitating key frame extraction.
Informative frames could better summarize the whole video where they reside, while how to quantify the information each frame contains is a hard-nut to crack.
Firstly, we calculate image entropy of each frame, and then map them into a two-dimensional coordinate space, as shown in  Figure \ref{Fig:key_frame_extra}(b).
Entropy is a nice way of representing the impurity or unpredictability of a set of data since it is dependent on the context in which the measurement is taken.
As for a single video frame, the gray-scale color/intensity distribution of this frame can be seen as $p = \{p_1, p_2, ... , p_n\}$.
For the image frames $f_i$, their image entropy can be defined as:
\begin{equation}
\label{Equ:entropy}
E(f_i) = -\sum _{j} p_\mathit{f_i}(j)logp_\mathit{f_i}(j),
\end{equation}
where $p_\mathit{f_i}(j)$ denotes the probability density function of frame $f_i$, which could be obtained by normalizing their histogram of gray-scale pixel intensities.
Next we map the value $E(f_i)$ to a two-dimensional coordinate space (the E($f_i$) vs. $i$ plot).

\subsubsection{Local Extreme Points.}
Secondly, we pick the local extreme points in the two-dimensional coordinate space, illustrated by Figure \ref{Fig:key_frame_extra}(c).
Local extreme points include the local maximum points and local minimum points.
Local maximum points can be calculated as follows:
\begin{equation}
\label{Equ:loc_max}
P_{max} =
\begin{cases}
E(f_i), ~~if~~E(f_i)> E(f_{i+1})~\&~E(f_i)> E(f_{i-1}). \\
remove, ~~else.
\end{cases}
\end{equation}
Local minimum points can also be calculated by the following formula:
\begin{equation}
\label{Equ:loc_min}
P_{min} =
\begin{cases}
E(f_i), ~~if~~E(f_{i+1})> E(f_i)~\&~E(f_{i-1})> E(f_i). \\
remove, ~~else.
\end{cases}
\end{equation}
where $i= 1, 2, ..., n$.
Therefore, local extreme points $P_\mathit{extreme}$ can be united by:
\begin{equation}
\label{Equ:extreme}
P_\mathit{extreme} = P_\mathit{max} \cup P_\mathit{min}.
\end{equation}

Local extreme points could further extract representative frames from the original video sequence.
This procedure could be viewed as finding local representatives to roughly describe the original sequence.

\subsubsection{Density Clustering.}
After obtaining the extreme points, as shown in Figure \ref{Fig:key_frame_extra}(c), we try to cluster these points into $N$ ($N$ is a pre-defined constant for all the datasets) categories, as, \{$1$, $2$\}, \{$3$, $4$, $5$, $6$, $7$\}, \{$8$\}, \{$9$\} and \{$10$\}.
The distribution of these extreme points have the characteristics, the cluster centers are surrounded by neighbors with a lower local density and that they are at a relatively large distance from any points with a higher local density.

Therefore, we adopt density clustering  \cite{rodriguez2014clustering}  to further cluster these extreme points $P_\mathit{extreme}$, as shown in Figure \ref{Fig:key_frame_extra}(d-f).
Density clustering could better catch the delicate structure of 2-D space where extreme points reside than traditional clustering strategies, e.g. K-means.
First, we search for a local density maximum point as a cluster center, and then spread the cluster label from high density to low-density points sequentially.
For each data point $P_k$, we compute two quantities: corresponding local density (neighborhood has a density, not the data point) $\rho_\mathit{P_k}$ and its distance $\delta_\mathit{P_k}$ from points of higher density.
Both these quantities depend only on the distances $d_\mathit{{P_k}{P_l}}$ between data points, which are assumed to satisfy the triangular inequality.
The local density $\rho_\mathit{P_k}$ of data point $P_k$ is defined as:
\begin{equation}
\label{Equ:cluster1}
\rho_\mathit{P_k} = \sum_\mathit{P_l} \chi(d_\mathit{{P_k}{P_l}} - d_c),
\end{equation}
where $\chi(x) = 1$ if $x < 0$ and $\chi(x)=0$ when otherwise, and $d_c$ is a cutoff distance.
Basically, $\rho_\mathit{P_k}$ is equal to the number of points that are closer than $d_c$ to point $P_k$.
The algorithm is sensitive only to the relative magnitude of $\rho_\mathit{P_k}$ in different points, which implies that, the results of the analysis are robust with respect to the choice of $d_c$ for large datasets.
A different way of defining $\rho_\mathit{P_k}$ as:
\begin{equation}
\label{Equ:cluster1_2}
\rho_\mathit{P_k} = \sum_\mathit{P_l} e^{-(\frac{d_\mathit{{P_k}{P_l}}}{d_c})^2}.
\end{equation}

$\delta_\mathit{P_k}$ is measured by finding the minimum distance between the point $P_k$ and any other point with higher density:
\begin{equation}
\label{Equ:cluster2}
\delta_\mathit{P_k} = \mathop{min}_\mathit{P_l: \rho_\mathit{P_l} >\rho_\mathit{P_k}}(d_\mathit{{P_k}{P_l}}),
\end{equation}
which uses a Gaussian kernel to calculate.
We can see from these two kernels, cutoff kernel is discrete value, while Gaussian kernel is a continuous value, which guarantees a smaller probability of conflict.

As we can see from  Figure \ref{Fig:key_frame_extra}(d), we calculate $\rho$ and $\delta$ using Formula (\ref{Equ:cluster1}) and (\ref{Equ:cluster2}).
Then select the number of clustering center $N$, namely, the $N$ largest $\delta$ values, e.g., in Figure \ref{Fig:key_frame_extra}(e), we select 5 cluster centers.
Figure \ref{Fig:key_frame_extra}(f) illustrates the final results of clustering, in which the points of 2, 5, 8, 9 and 10 are the clustering centers, therefore, the corresponding x-coordinates (the 2, 9, 14, 20 and 26 frames, shown in figure \ref{Fig:key_frame_extra}(g)) are the key frames $S_\mathit{Keyframes}$ in the original video $V$ (shown in Figure \ref{Fig:key_frame_extra}(a)).
The pipeline of the proposed key frames extraction method is summarized in Algorithm \ref{Alg:key_frame_framwork}.
Note that the proposed density clustering cannot handle the situation where the entropy of the video sequence is monotone increasing or decreasing since we need to select the local extreme points.
While in our experiments, we observer that there is no one video sequence which frame entropy is monotone increasing or decreasing all the time, it means we can always obtain the local extreme points.

\begin{algorithm}[!t] \small
	\caption{The proposed key frames extraction method.}
	\label{Alg:key_frame_framwork}
	\begin{algorithmic}[1] 
		\REQUIRE ~~\\
		The original hand gesture video $V$, as shown in Figure \ref{Fig:key_frame_extra}(a) and the number of key frames $N$ ($N$ is a pre-defined constant for all the datasets).
		\ENSURE ~~\\ 
		The key frames $S_\mathit{Keyframes}$ in original video $V$, as shown in Figure \ref{Fig:key_frame_extra}(g).
		\STATE Calculate image entropy $E(f_i)$ of each frame in $V$ using Formula \ref{Equ:entropy};
		\STATE Map $E(f_i)$ to a two-dimensional coordinate space;
		\STATE Find local maximum points $P_\mathit{max}$ in the two-dimensional coordinate space using Formula (\ref{Equ:loc_max});
		\STATE Find local minimum points $P_\mathit{min}$ in the two-dimensional coordinate space using Formula (\ref{Equ:loc_min});
		\STATE Obtain $P_\mathit{extreme}$ by uniting the local maximum points $P_\mathit{max}$ and local minimum points $P_\mathit{min}$ using Formula (\ref{Equ:extreme});
		\STATE Calculate $\rho$ for each point in $P_\mathit{extreme}$ using Formula (\ref{Equ:cluster1}) or (\ref{Equ:cluster1_2});
		\STATE Calculate $\delta$ for each point in $P_\mathit{extreme}$ using Formula (\ref{Equ:cluster2});
		\STATE Draw decision graph like Figure \ref{Fig:key_frame_extra}(d);
		\STATE Choose the $N$ largest $\delta$ values as the clustering centers, as shown in Figure \ref{Fig:key_frame_extra}(e);
		\STATE The corresponding x-coordinate of $N$ clustering centers are the key frames $S_\mathit{Keyframes}$.
		\RETURN $S_\mathit{Keyframes}$.
	\end{algorithmic}
\end{algorithm}

\subsection{Feature Fusion Strategy}
In view of each key frame and the relationship between each key frame, for better representing each key frame sequence, we not only try to describe each frame of key frame, but also the variation between the key frames.
That is, we not only extract the most representative frames and map them into a simple 2-D space, but also describe how these frames move in this space.
We believe this two-phase strategy could set up a ``holographic'' description of each hand gesture sequence.
Hadid and Pietik{\"a}inen \cite{hadid2009combining} also demonstrate that excellent results can be obtained combining appearance feature and motion feature for dynamic video analysis.
Jain et al. \cite{jain2017fusionseg} propose a two-stream fully convolutional neural network which fuses together motion and appearance in a
unified framework, and substantially improve the state-of-the-art results for segmenting unseen objects in videos.
Xu et al. \cite{xu2017video}  consider exploiting the appearance and motion information resided in the video with a attention
mechanism for the image question answering task.
For this purpose, we propose a feature fusion strategy to capture these two phases: appearance based approach can only be applied to each frame, which represents the differences of space merely; while, motion based method can describe the evolution with the time.
Thus we combine appearance and motion feature for better describe image sequence.
Meanwhile, to better weight these two feature, we also introduce an efficient strategy to balance them.
Tang et al.  \cite{tang2018gesturegan} also propose a feature fusion method which fuses features extracted from different sleeves to boost the recognition performance.

\begin{figure*}[!tbp]
	\begin{centering}
		\includegraphics[width=1\linewidth]{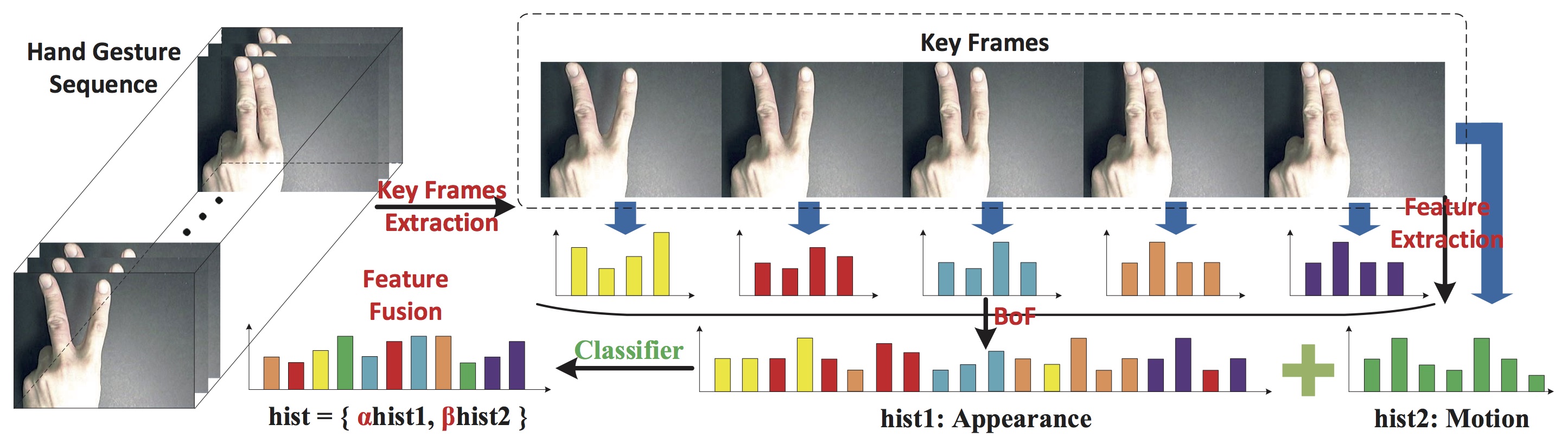}
		\caption{The framework of the proposed feature extraction and fusion methods.}
		\label{Figure:feature_fusion}
	\end{centering}
\end{figure*}

Figure \ref{Figure:feature_fusion} shows the whole proposed feature fusion procedure for the obtained key frames.
After extracting key frames, we take the key frames sequence in place of the original sequence.
We begin by extracting key frames from the original sequence, and then extract appearance and motion features (\textbf{hist1} and \textbf{hist2} in Figure \ref{Figure:feature_fusion}) from the key frames sequence, respectively.
For further increase the importance of the useful feature, we add weights to appearance and motion features.
By feeding \textbf{hist1} and \textbf{hist2} to the SVM classifier separately, we obtain two classification accuracy $R = \{R_a, R_m\}$.
Based on the assumption that the higher the rate is, the better representation becomes, we compute the weights as follows:
\begin{equation}
T = \frac{R - min(R)}{(100 - min(R))/10}.
\label{equ:t}
\end{equation}
Finally, considering that the weight of the lowest rate is 1, the other weights can be obtained according to a linear relationship of their differences to that with the lowest rate.
The final step is written as:
\begin{equation}
\begin{aligned}
T1 &= round(T)\\
T2 &= \frac{T \times ((max(T1)-1))}{max(T)} + 1 \\
W  &= round(T2)
\end{aligned}
\label{equ:f}
\end{equation}
in which $W= \{\alpha, \beta\}$ is the weight vector corresponding to \textbf{hist1} and \textbf{hist2}.

There are many existing descriptors for us to extract \textbf{hist1} and \textbf{hist2}.
In other words, the fusion strategy does not depend on specific descriptors, which guarantees its great potential in applications.
In term of \textbf{hist1}, we can use Gist \cite{oliva2001modeling}, rootSIFT \cite{arandjelovic2012three}, HSV, SURF \cite{bay2006surf}, HOG \cite{dalal2005histograms}, LBP \cite{ojala2002multiresolution} or its variation CLBP \cite{guo2010completed} to extract appearance cue of each image.
As for \textbf{hist2}, LBP-TOP, VLBP \cite{zhao2007dynamic} and SIFT 3D \cite{scovanner20073} are used to extract motion cues from the whole key frames sequence.
We also use Bag-of-Feature (BoF) \cite{liu2016sequential,dardas2011real} to represent these appearance and motion cues.
At the end of the procedure, we concatenate weighted \textbf{hist1} and \textbf{hist2} to obtain the final representation \textbf{hist} (as shown in Figure \ref{Figure:feature_fusion}).

\subsection{Hand Gesture Recognition Framework}
The hand gesture recognition framework based on key frames and feature fusion is composed of two stages, training and testing, which is summarized in Algorithm \ref{Alg:whole_framwork}.
In the training stage, we first extract key frames $S_\mathit{Keyframes}$ using Algorithm \ref{Alg:key_frame_framwork} (step \ref{Code:key_frame_extrat}).
Then we extract appearance features from each key frame using descriptors such as SURF, LBP, etc.
After obtaining the appearance features, we employ BoF \cite{liu2016sequential,dardas2011real} to represent these features for ${hist1}^q$ (step \ref{Code:app_extrat}).
Next we use LBP-TOP, VLBP or SIFT 3D to extract motion features from the whole key frames sequence, producing the corresponding histogram ${hist2}^q$ (step \ref{Code:mon_extrat}).
After that, ${hist1}^q$ and ${hist2}^q$ are fed to separate classifiers to obtain $R= \{R_a, R_m\}$ (step \ref{Code:r}).
And then, we calculate $\alpha$ and $\beta$ by Formula (\ref{equ:t}) and (\ref{equ:f}) (step \ref{Code:alp_bet}).
Then the training histogram ${hist}^q$ is constructed from $\alpha{hist1}^q$ and $\beta{hist2}^q$ (step \ref{Code:concatenated}).
In the end of the iteration, we obtain the training representation vector $hist$.
Then $hist$ and corresponding labels $L_{label}$ are fed to a SVM classifier (step \ref{Code:class}).
During the testing stage, testing hand gesture representation is obtained in the same way as the training stage (step \ref{Code:test}).
Thereby the trained SVM classifier is used to predict the gesture label $t_{label}$ (step \ref{Code:predict}).

\begin{algorithm}[!t] \small
	\caption{The proposed hand gesture recognition framework.}
	\label{Alg:whole_framwork}
	\begin{algorithmic}[1]
		\REQUIRE ~~\\
		$L$ hand gesture videos for training, as shown in Figure \ref{Fig:key_frame_extra}(a), corresponds to the gesture labels $L_\mathit{label}$; \\
		Testing hand gesture video $t$.
		\ENSURE ~~\\
		The hand gesture label $t_\mathit{label}$.
		\STATE TRAINING STAGE:
		\FOR{$q = 1$ to $L$}
		\STATE $S_\mathit{Keyframes}^{q}$ $\gets$ Algorithm \ref{Alg:key_frame_framwork}; \label{Code:key_frame_extrat}
		\STATE ${hist1}^q$ $\gets$ (SURF or GIST, etc) $\cup$ BoF; \label{Code:app_extrat}
		\STATE ${hist2}^q$ $\gets$ (VLBP or LBP-TOP or SIFT 3D, etc) $\cup$ BoF; \label{Code:mon_extrat}
		\STATE $R= \{R_a, R_m\}$ $\gets$ training a classifier using ${hist1}^q$ and ${hist2}^q$ ; \label{Code:r}
		\STATE $\alpha$ and $\beta$ $\gets$ using by Formula (\ref{equ:t}) and (\ref{equ:f}); \label{Code:alp_bet}
		\STATE ${hist}^q$ $\gets$ \{$\alpha$hist1, $\beta$hist2\}; \label{Code:concatenated}
		\ENDFOR
		\STATE Classifier $\gets$ ${hist}$ $\cup$ $L_\mathit{label}$; \label{Code:class}
		\STATE TESTING STAGE:
		\STATE Obtain hand gesture representation ${hist}^t$ for testing $t$ using the same method as the training stage; \label{Code:test}
		\STATE Obtain $t_\mathit{label}$ by the classifier after calculation; \label{Code:predict}
		\RETURN $t_\mathit{label}$.
	\end{algorithmic}
\end{algorithm}

\section{Experiments and Analysis}
\label{sec:exper}

\subsection{Datasets and Settings}
To evaluate the effectiveness of the proposed method, we conduct experiments on two publicly available datasets (Cambridge 
\cite{kim2007tensor} and Northwestern University Hand Gesture datasets \cite{shen2012dynamic}) and two collected datasets (HandGesture and Action3D hand gesture datasets, both will be released after paper accepted).
Some characteristics of these datasets are listed in Table \ref{Tab:datasets}.

\textbf{Cambridge Hand Gesture} dataset is a commonly used benchmark gesture data set with 900 video clips of 9 hand gesture classes defined by 3 primitive hand shapes (i.e., flat, spread, V-shape) and 3 primitive motions (i.e., leftward, rightward, contract).
For each class, it includes 100 sequences captured with 5 different illuminations, 10 arbitrary motions and 2 subjects.
Each sequence is recorded in front of a fixed camera having coarsely isolated gestures in spatial and temporal dimensions.

\textbf{Northwestern University Hand Gesture} dataset is a more diverse data set which contains 10 categories of dynamic hand gestures in total: move right, move left, rotate up, rotate down, move downright, move right-down, clockwise circle, counterclockwise circle, ``Z'' and cross.
This dataset is performed by 15 subjects and each subject contributes 70 sequences of these ten categories with seven postures (i.e., Fist, Fingers extended, ``OK'', Index, Side Hand, Side Index and Thumb).

These two datasets mentioned above are both with clear backgrounds, and sequences snipped tightly around the gestures.
However, how well will this method work on videos from ``the wild'' with significant clutter, extraneous motion, continuous running video without pre-snipping?
To validate our approach, we introduce two new datasets, called HandGesture and Action3D datasets.

\textbf{HandGesture} data set consists of 132 video sequences of $640$ by $360$ resolution, each of which recorded from a different subject (7 males and 4 females) with 12 different gestures (``0''-``9'', ``NO'' and ``OK'').

We also acquired \textbf{Action3D} dataset which consisting of 1620 image sequences of 6 hand gesture classes (box, high wave, horizontal wave, curl, circle and hand up), which are defined by 2 different hands (right and left hand) and 5 situations (sit, stand, with a pillow, with a laptop and with a person).
Each class contains 270 image sequences (5 different situations $\times$ 2 different hands $\times$ 3 times $\times$ 9 subjects).
Each sequence was recorded in front of a fixed camera having roughly isolated gestures in space and time.
All video sequences were uniformly resized into $320 \times 240$ in our method.

\begin{table*}[!tbp]
	\centering
	\caption{Characteristics of the datasets used in our hand gesture recognition experiments.}
	\begin{tabular}{c|c|c|c|c|c} \hline
		Dataset                        & \# categories & \# videos & \# training  & \# validation & \# testing \\ \hline
		\emph{Cambridge}       & 9             & 900       & 450          & 225            & 225           \\ \hline
		\emph{Northwestern}  & 10            & 1,050      & 550          & 250            & 250         \\ \hline
		\emph{HandGesture}   & 12            & 132       & 66           & 33             & 33               \\ \hline
		\emph{Action3D}         & 6             & 1,620      & 810          & 405            & 405         \\ \hline
	\end{tabular}
	\label{Tab:datasets}
\end{table*}

\subsection{Parameter Analysis}
\begin{figure*}[!t]
	\begin{centering}
		\centering
		\setcounter{subfigure}{0}
		\subfigure[]  {\includegraphics[width=2.5in]{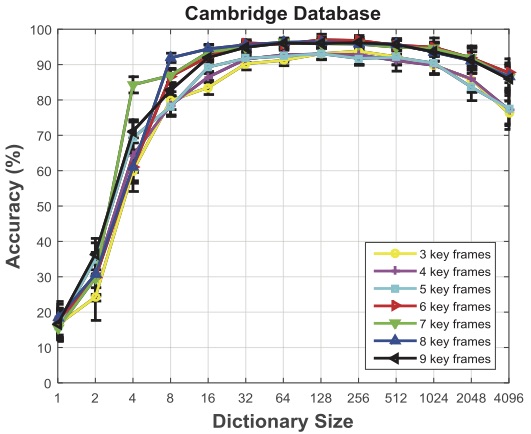}}
		\subfigure[]  {\includegraphics[width=2.5in]{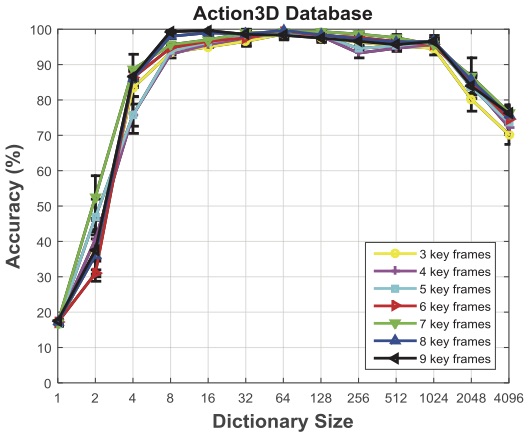}} \\
		\subfigure[]  {\includegraphics[width=2.5in]{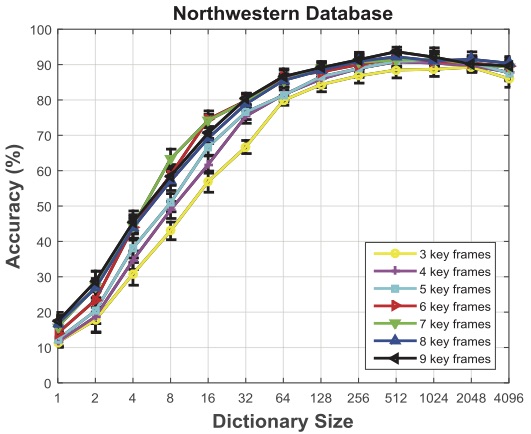}}
		\subfigure[]  {\includegraphics[width=2.5in]{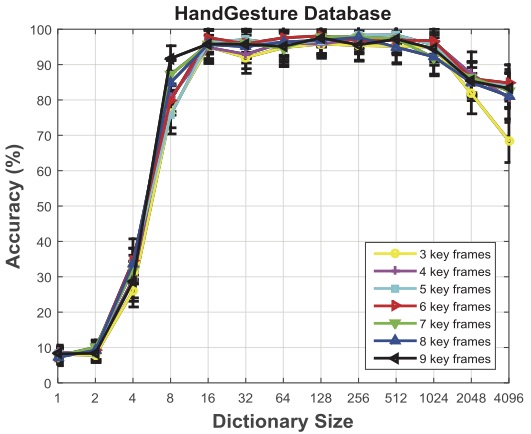}} \\
		\caption{Parameters $N$ and $D$ selection on the four datasets.}
		\label{Fig:papameter}
	\end{centering}
\end{figure*}

Two parameters are involved in our framework: the number of key frames $N$ and the dictionary number $D$ in BoF.
Firstly, we extract $N = 3, 4, ..., 9$ key frames from the original video, respectively.
And then extract SURF features from each key frame.
Every key point detected by SURF provides a 64-D vector describing the texture of it.
Finally, we adopt BoF to represent each key frame with dictionary $D = 1, 2, 4, ..., 4096$, respectively.
The number of training set, validation set and testing set please refer to Table \ref{Tab:datasets}.
We repeat all the experiments 20 times with different random spits of the training and testing samples to obtain reliable results.
The final classification accuracy is reported as the average of each run.
Figure \ref{Fig:papameter} presents the accuracy results on the four datasets.
From Figure \ref{Fig:papameter} (a) and (c), the accuracy first rises to the peak when $D = 64$ and then drops after reaching the peak.
However, as shown in Figure \ref{Fig:papameter} (b) and (d), the accuracy reaching the peak when $D = 16$.
Thus, we set $D = 64$ on the Cambridge and Northwestern datasets, and $D = 16$ on our two proposed datasets.
It is observe that the more key frames we have, the more time will be consumed.
Thus, to balance accuracy and efficiency, we set $N = 5$ on all the four datasets in the following experiments.

\subsection{Experiment Evaluation}
To evaluate the necessity and efficiency of the proposed strategy, we test it in multi-aspect: (1) necessity of key frames extraction; (2) different kernel tricks; (3) different fusion strategies; (4) different clustering methods; (5) performance comparisons with the state-of-the-art; (6) efficiency.
(1)-(4) demonstrate the rationality and validity of the methods.
(5) compares the proposed method with others.
(6) shows its efficiency.

\noindent (1) Comparison with Different Key Frames Extraction Methods.
We discuss whether our key frames method is necessary or not here.
For the Cambridge and Action3D dataset, we only extract LBP-TOP, and then concatenate the three orthogonal planes of LBP-TOP.
For the Northwestern and HandGesture dataset, 200 points are randomly selected from each video using SIFT 3D.
As we can see from Table \ref{Tab:Comparison_key_or_not}, our approach outperforms the other four methods on both accuracy and efficiency, thereby our approach is not only theoretical improvement, but also has an empirical advantage.

\begin{table*}[!t]
	\centering
	\caption{Comparison between different key frames extraction methods on the Cambridge, Northwestern, HandGesture and Action3D datasets.}
	\begin{tabular}{c|cc|cc}
		\hline
		\multirow{2}{*}{Method}    & \multicolumn{2}{|c|}{\emph{Cambridge}} & \multicolumn{2}{|c}{\emph{Northwestern}}  \\ \cline{2-5}
		& Accuracy                     & Time(s)  & Accuracy          & Time(s)  \\ \hline
		Original Sequence                                                & $35.26\% \pm 3.15\%$  & 20,803   & $21.65\% \pm 1.23\%$  & 108,029  \\
		5  evenly-spaced frames (in time)                        & $56.13\% \pm 5.46\%$  & 1,189      & $58.79\% \pm 2.64\%$  & 26,303   \\
		Zhao and Elgammal \cite{zhao2008information}  & $58.14\% \pm 3.36\%$      & 1,432    & $61.45\% \pm 3.45\%$  & 27,789   \\
		Carlsson and Sullivan \cite{carlsson2001action}   & $50.57\% \pm 4.78\%$      & 1,631    & $51.27\% \pm 3.86\%$  & 29,568   \\ \hline
		Ours key frames method                                      & $60.78\% \pm 2.21\%$       & 1,152    & $64.24\% \pm 2.15\%$  & 25,214   \\ \hline \hline
		
		\multirow{2}{*}{Method}    & \multicolumn{2}{|c|}{\emph{HandGesture}} & \multicolumn{2}{|c}{\emph{Action3D}} \\ \cline{2-5}
		& Accuracy                    & Time(s) &  Accuracy           & Time(s) \\ \hline
		Original Sequence                                                & $42.54\% \pm 4.61\%$ & 8,549   &  $34.56\% \pm 2.65\%$ & 284,489 \\
		5  evenly-spaced frames (in time)                         & $58.32\% \pm 3.88\%$ & 1,689   &  $52.13\% \pm 2.31\%$ & 18,430 \\
		Zhao and Elgammal \cite{zhao2008information}   & $60.45\% \pm 4.56\%$ & 1,895   &  $54.56\% \pm 1.97\%$ & 20,143  \\
		Carlsson and Sullivan  \cite{carlsson2001action}   & $50.78\% \pm 4.06\%$ & 2,154   &  $46.34\% \pm 2.78\%$ & 23,768  \\ \hline
		Ours key frames method                                       & $65.18\% \pm 3.62\%$ & 1,645   &  $56.13\% \pm 1.89\%$ & 16,294 \\ \hline
	\end{tabular}
	\label{Tab:Comparison_key_or_not}
\end{table*}

\noindent (2) Gaussian Kernel vs. Cutoff Kernel.
We also compare the Gaussian and cutoff kernel.
We adopt  SURF to extract feature from each key frame.
Comparison results are shown in the Table \ref{Tab:kernel}.
As we can observe that there is small different between using the Gaussian and cutoff kernel.

\begin{table}[!htbp]
	\centering
	\caption{Comparison between Gaussian kernel and cutoff kernel on the Cambridge, Northwestern, HandGesture and Action3D datasets.}
	\begin{tabular}{c|c|c} \hline
		Kernel                          & Gaussian Kernel          & Cutoff Kernel\\ \hline
		\emph{Cambridge}       & 92.37\% $\pm$ 1.67\% & 90.33\% $\pm$ 2.78\%   \\ \hline
		\emph{Northwestern}  & 81.31\% $\pm$ 1.49\%  & 80.25\% $\pm$ 1.86\%   \\ \hline
		\emph{HandGesture}   & 96.32\% $\pm$ 3.35\% & 94.54\% $\pm$ 2.78\%    \\ \hline
		\emph{Action3D}         & 96.26\% $\pm$ 1.39\% & 93.65\% $\pm$ 1.23\%   \\ \hline
	\end{tabular}
	\label{Tab:kernel}
\end{table}

\noindent (3) Comparison with Different Feature Fusion Strategies.
We demonstrate that combining appearance (hand posture) and motion (the way hand is moving) boosts hand gesture recognition task here.
Moreover, we also compare different schemes based on appearance and motion, respectively.
For feature fusion, we set $\alpha$ and $\beta$ to 8 and 1, respectively.
As shown in Table \ref{Tab:fusion_stra}, fusion strategies are much better than appearance or motion based methods, which demonstrates the necessity of our feature fusion strategy.
Motion-based method achieve the worst results, which can illustrate that in our task spatial cues is more important than the temporal cues.
The spatial cues represents/extracts the difference between different gesture classes, also called inter-class differences.
While the temporal cues captures the difference among different frames in the same gesture sequences, also called intra-class differences.
Inter-class differences are always greater than intra-class differences, which means the spatial cues can represent more discriminative feature than the temporal cues.
In our task, we observe that the differences between different types of gestures are much greater than the differences between the same gesture sequence, which means the spatial cues is more discriminative than the temporal cues.
However, if hand gesture moves fast and change hugely in one sequence, the temporal cues could be more important.

\begin{table*}[!htbp]
	\centering
	\caption{Comparison with different features fusion strategies on the Cambridge, Northwestern, HandGesture and Action3D datasets. $N$ and $D$ denote the number of key frames and the dictionary number.}
	\resizebox{\linewidth}{!}{%
		\begin{tabular}{c|c|c|c|c|c} \hline
			Appearance-based       & Dimension & \emph{Cambridge}      & \emph{Northwestern} & \emph{HandGesture} & \emph{Action3D}  \\ \hline
			SURF                            & $N*D$      & $92.37\% \pm 1.67\%$ & $81.31\% \pm 1.49\%$ & $96.32\% \pm 3.35\%$  & $96.26\% \pm 1.39\%$ \\
			GIST                             & $N*D$       & $88.15\% \pm 1.65\%$ & $78.41\% \pm 1.91\%$  & $92.64\% \pm 2.89\%$  & $91.23\% \pm 1.84\%$  \\ \hline \hline
			
			Motion-based               & -              & \emph{Cambridge}       & \emph{Northwestern} & \emph{HandGesture} & \emph{Action3D}  \\ \hline
			LBP-TOP                       & 177          & $60.78\% \pm 2.21\%$ & $51.36\% \pm 2.16\%$ & $60.84\% \pm 2.61\%$   & $56.13\% \pm 1.89\%$   \\
			VLBP                             & 16,386     & $50.36\% \pm 3.56\%$ & $42.11\% \pm 3.04\%$ & $49.78\% \pm 3.51\%$   & $44.26\% \pm 4.23\%$ \\
			SIFT 3D                         &  $N*D$    & $68.94\% \pm 4.81\%$ & $64.24\% \pm 2.15\%$  & $65.18\% \pm 3.62\%$   & $62.04\% \pm 2.89\%$ \\ \hline \hline
			
			Appearance + Motion    & -                               & \emph{Cambridge}        & \emph{Northwestern}       & \emph{HandGesture}       & \emph{Action3D}  \\ \hline
			SURF + LBP-TOP           & $N*D$+177       & $95.75\% \pm 0.79\%$      & $93.54\% \pm 1.36\%$        & $97.25\% \pm 0.79\%$       & $98.53\% \pm 1.31\%$ \\
			SURF + VLBP                 & $N*D$+16,386 & $92.52\% \pm 1.27\%$      & $91.22\% \pm 0.95\%$        & $96.82\% \pm 0.95\%$       & $97.21\% \pm 0.94\%$  \\
			SURF + SIFT 3D             & $2*N*D$            & \bm{$98.23\% \pm 0.84\%$} & \bm{$96.89\% \pm 1.08\%$}   & \bm{$99.21\% \pm 0.88\%$}  & \bm{$98.98\% \pm 0.65\%$ } \\
			GIST + LBP-TOP            & $N*D$+177      & $91.87\% \pm 1.65\%$      & $86.26\% \pm 0.94\%$        & $93.56\% \pm 1.35\%$       & $93.11\% \pm 0.89\%$  \\
			GIST + VLBP                  & $N*D$+16,386 & $90.56\% \pm 0.87\%$      & $82.87\% \pm 1.84\%$        & $92.88\% \pm 1.21\%$       & $92.63\% \pm 0.64\%$   \\
			GIST + SIFT 3D              & $2*N*D$           & $93.52\% \pm 0.63\%$      & $88.54\% \pm 1.62\%$        & $94.16\% \pm 0.67\%$       & $94.21\% \pm 0.61\%$   \\ \hline
	\end{tabular}}
	\label{Tab:fusion_stra}
\end{table*}

\noindent (4) Comparison with Different Clustering Methods.
We also compare different clustering methods for the key frames extraction.
As shown in Table \ref{Tab:clustering}, we can see that density clustering is much better than K-means, OPTICS \cite{ankerst1999optics} and DBSCAN \cite{ester1996density}.

\begin{table*}[!htbp]
	\centering
	\caption{Comparison with different clustering methods on the Cambridge, Northwestern, HandGesture and Action3D datasets.}
	\begin{tabular}{c|c|c|c|c} \hline
		Clustering Method     & OPTICS \cite{ankerst1999optics} & DBSCAN \cite{ester1996density} & K-means  & Density Clustering \cite{rodriguez2014clustering} \\ \hline
		\emph{Cambridge}      & $88.15\% \pm 1.51\%$ & $90.34\% \pm 1.78\%$ & $86.26\% \pm 2.51\%$ & $98.23\% \pm 0.84\% $    \\ \hline
		\emph{Northwestern} & $86.34\% \pm 2.45\%$ & $88.35\% \pm 1.67\%$ & $83.65\% \pm 1.06\%$ & $96.89\% \pm 1.08\% $     \\ \hline
		\emph{HandGesture}  & $84.56\% \pm 1.89\%$ & $85.98\% \pm 1.76\%$ & $84.69\% \pm 1.98\%$ & $99.21\% \pm 0.88\% $    \\ \hline
		\emph{Action3D}        & $83.56\% \pm 1.56\%$ & $87.43\% \pm 1.63\%$& $82.36\% \pm 1.46\%$ & $98.98\% \pm 0.65\% $    \\ \hline
	\end{tabular}
	\label{Tab:clustering}
\end{table*}

\noindent (5) Comparison with State-of-the-Arts.
For the Cambridge and Northwestern datasets, we compare our results with the state-of-the-art methods in Tables \ref{Tab:cambridge} and \ref{Tab:northwestern}.
We achieve $98.23\% \pm 0.84\%$ and $96.89\% \pm 1.08\%$ recognition accuracy on the Cambridge and Northwestern dataset, both of which exceed the other baseline methods.

\begin{table*}[!tbp]
	\caption{Comparison with the state-of-the-art methods on the Cambridge dataset.}
	\centering
	\begin{tabular}{c|c|c}
		\hline
		\emph{Cambridge}                             & Methods                                 & Accuracy \\ \hline 
		Wong and Cipolla \cite{wong2005real}         & Sparse Bayesian Classifier              & 44\%  \\ 
		Niebles et al. \cite{niebles2008unsupervised}& Spatial-Temporal Words                  & 67\% \\
		Kim et al. \cite{kim2007tensor}              & Tensor Canonical Correlation Analysis   & 82\% \\
		Kim and Cipolla \cite{kim2009canonical}      & Canonical Correlation Analysis          & 82\% \\
		Liu and Shao \cite{liu2013synthesis}         & Genetic Programming                     & 85\% \\
		Lui et al. \cite{lui2010action}              & High Order Singular Value Decomposition & 88\% \\
		Lui and Beveridge \cite{lui2011tangent}      & Tangent Bundle                          & 91\% \\
		Wong et al. \cite{wong2007learning}          & Probabilistic Latent Semantic Analysis  & 91.47\% \\
		Sanin et al. \cite{sanin2013spatio}          & Spatio-Temporal Covariance Descriptors  & 93\% \\
		Baraldi et al. \cite{baraldi2014gesture}     & Dense Trajectories + Hand Segmentation  & 94\% \\
		Zhao and Elgammal \cite{zhao2008information} & Information Theoretic                   & 96.22\% \\ \hline
		Ours                                   & Key Frames + Feature Fusion &\bm{$98.23\% \pm 0.84\%$} \\ \hline
	\end{tabular}
	\label{Tab:cambridge}
\end{table*}

\begin{table*}[!tbp]
	\caption{Comparison between the state-of-the-art methods and our method on the Northwestern University dataset.}
	\centering
	\begin{tabular}{c|c|c}
		\hline
		\emph{Northwestern}                  & Methods                     & Accuracy \\ \hline 
		Liu and Shao \cite{liu2013synthesis} & Genetic Programming         & 96.1\%  \\
		Shen et al. \cite{shen2012dynamic}   & Motion Divergence fields    & 95.8\%  \\ \hline
		Our method                           & Key Frames + Feature Fusion & \bm{$96.89\% \pm 1.08\%$}  \\ \hline
	\end{tabular}
	\label{Tab:northwestern}
\end{table*}

\begin{table*}[!htbp]
	\centering
	\caption{Computation time for classifying a test sequence on the Cambridge, Northwestern, HandGesture and Action3D datasets.}
	\begin{tabular}{c|c|c|c|c} \hline
		Time                                        & \emph{Cambridge} & \emph{Northwestern} & \emph{HandGesture} & \emph{Action3D} \\ \hline
		Entropy Calculation                 &  0.93s          & 0.84s                &   3.21s            & 0.75s \\
		Density Clustering                   &  0.31s          & 0.34s                 &   0.43s            & 0.38s \\
		Feature Extraction                   &  3.07s          & 9.71s                  &  9.42s             & 3.13s \\  
		SVM Classification                  &  0.60 ms       & 0.51ms               & 0.46ms          & 0.65 ms \\  \hline \hline
		Our Full Model                        &  4.31s           & 10.89s                & 13.06s            & 4.26s  \\ \hline 
		Liu and Shao \cite{liu2013synthesis} & 6.45s & 13.32s                & 15.32s              & 6.43s \\ \hline
		Zhao and Elgammal \cite{zhao2008information} & 5.34s & 11.78s       & 14.98s             &  5.21s \\ \hline
	\end{tabular}
	\label{Tab:efficiency}
\end{table*}

\noindent (6) Efficiency.
Finally, We investigate our approach in terms of computation time of classifying one test video.
We run our experiment on a 3.40-GHz i7-3770 CPU with 8 GB memory.
The proposed method is implemented using Matlab.
Code and datasets will release after paper accepted.
As we can see from Table \ref{Tab:efficiency} that the time of classifying one test sequence is 4.31s, 10.89s, 13.06s and 4.26s for the Cambridge, Northwestern, HandGesture and Action3D datasets.
We observe that the proposed key frame extraction methods including entropy calculation and density clustering can be finished within around 1s on the Cambridge, Northwestern and Action3D datasets.
While for the HandGesture dataset which contains about 200 frames in a single video, it only cost about 3s per video. 
We also note that the most time-consuming part is feature extraction, and we have two solutions to improve it, (i) we can reduce the size of images, we note that Cambridge and Action3D only consume about 4s, while Northwestern and HandGesture cost about 11s and 13s respectively.
The reason is that the image size of Cambridge and Action3D is $320\times240$, while the image size of Northwestern and HandGesture is $640\times480$; (ii) We can further reduce the time for feature extraction by using a GPU like most deep learning methods do.
Moreover, we also compare two methods ( \cite{zhao2008information} and \cite{liu2013synthesis}  currently achieve best recognition results on the Cambridge and Northwestern datasets, respectively.) on the time for classifying a test sequence on the Cambridge, Northwestern, HandGesture and Action3D datasets.
We re-implement both methods with the same running settings for fair comparison, including hardware platform  and programming language.
The results are shown in Table \ref{Tab:efficiency}, and we can see that the proposed method achieve better results than both methods.

\section{Conclusion}
\label{sec:concl}
In order to build a fast and robust gesture recognition system, in this paper, we present a novel key frames extraction method and feature fusion strategy.
Considering the speed of recognition, we propose a new key frames extraction method based on image entropy and density clustering, which can greatly reduce the redundant information of original video.
Moreover, we further propose an efficient feature fusion strategy which combines appearance and motion cues for robust hand gesture recognition.
Experimental results show that the proposed approach outperforms the state-of-the-art methods on the Cambridge ($98.23\% \pm 0.84\%$) and Northwestern ($96.89\% \pm 1.08\%$) datasets.
For evaluate our method on videos from ``the wild'' with significant clutter, extraneous motion and no pre-snipping, we introduce two new datasets, namely HandGesture and Action3D.
We achieve accuracy of~$99.21\% \pm 0.88\%$ and $98.98\% \pm 0.65\%$ on the HandGesture and Action3D datasets, respectively.
From the respect of the recognition speed, we also achieve better results than the state-of-the-art approaches for classifying one test sequence on the Cambridge, Northwestern, HandGesture and Action3D datasets.

\section*{Acknowledgments}
This work is partially supported by National Natural Science Foundation of China (NSFC, U1613209),  Shenzhen Key Laboratory for Intelligent Multimedia and Virtual Reality (ZDSYS201703031405467), Scientific Research Project of Shenzhen City (JCYJ20170306164738129).

\section*{References}
\bibliographystyle{elsarticle-num}
\bibliography{rough}

\end{document}